\documentclass{article}
\usepackage[accepted]{icml2025}
\makeatletter
\@ifpackageloaded{algorithmic}{%

}{}
\makeatother

\usepackage{algorithm}

\usepackage{algpseudocode}  

\usepackage{amsmath}
\usepackage{amssymb}
\usepackage{booktabs}
\usepackage{enumitem}
\usepackage{graphicx}
\usepackage{url}
\usepackage{placeins}
\usepackage{caption}  
\usepackage{hyperref}

\icmltitlerunning{Lateral Tree-of-Thoughts Surpasses ToT by Incorporating…}
\begin{document}

\twocolumn[
\icmltitle{Lateral~Tree\mbox{-}of\mbox{-}Thoughts Surpasses ToT by Incorporating Logically\mbox{-}Consistent, Low\mbox{-}Utility Candidates}

\begin{icmlauthorlist}
\icmlauthor{Abhinav Madahar}{ind}
\end{icmlauthorlist}
\icmlaffiliation{ind}{Independent Computer Scientist}
\icmlcorrespondingauthor{Abhinav Madahar}{abhinavmadahar@gmail.com}

\vskip 0.3in
]

\begin{abstract}
Modern deployments increasingly budget \emph{large test-time compute}---thousands of tokens or many node expansions---to improve reliability. When \emph{structured search} (e.g., ToT/MCTS-style controllers) is run under such frontier-like conditions, two effects intensify: \emph{breadth saturation}, where additional samples at a node mostly yield near-duplicates so width stops growing; and \emph{myopia}, where early, noisy utility undervalues branches whose payoff appears only after a few more steps, so they are pruned too soon. We introduce \textbf{Lateral Tree-of-Thoughts (LToT)}, a search-time controller that explicitly separates the frontier into \emph{mainlines}---\emph{high-utility} candidates used for exploitation---and \emph{laterals}---\emph{logically consistent, initially low-utility} candidates that merit short, cheap probes before judgment. LToT explores laterals via \emph{Lateral Racing with Short-Circuit (LR--SC)}, a budgeted race that spreads tiny probes across a very wide lateral set, culls aggressively, and immediately promotes a branch once it demonstrably clears the mainline bar; mainlines are kept intentionally narrow so surplus compute is invested where width is cheap. This turns large budgets into principled diversity while preserving promotion discipline. Let $N_{0}$ denote the \emph{initial lateral width} (the number of laterals admitted to the race) and $\eta>1$ the \emph{culling factor} between rungs. We show a \emph{pseudolinear lateral cost} $\Theta(N_{0}\log_{\eta}N_{0})$ with logarithmically many rungs. Pseudolinearity matters because it allows lateral width to scale almost linearly in cost, so increasing budgets buy \emph{coverage} rather than redundant deepening. Under \emph{equal compute}, we evaluate on GSM-Hard/Plus, MATH-500, HumanEval, MBPP-lite, and Game-of-24, reporting Success@1/Pass@1, width scaling, \emph{time-to-first-verified} solution, and \emph{false-promotion} rates. Across math, code, and ToT-style puzzles, LToT improves or matches accuracy while \emph{reducing expansions-to-first-hit}, converting surplus test-time compute into \emph{productive breadth} without sacrificing selectivity.
\end{abstract}

\section{Introduction}\label{section:introduction}

Modern language models (LMs) are increasingly deployed in \emph{compute--rich} inference regimes: users and systems budget thousands of tokens or many node expansions per query in return for reliability.
The dominant recipe for using this budget is structured search, most commonly Tree--of--Thoughts (ToT) over partial solutions \citep{yao2023tot}, layered on top of stepwise prompting \citep{wei2022cot,wang2022selfconsistency,kojima2022zeroshotcot}.

We argue that inference--time controllers should treat \emph{logically consistent, low--utility} candidates as assets, not waste.
The architectural move is to separate \emph{consistency/continuability} from \emph{utility}, and to invest cheap budget into short \emph{predictive continuations} of many consistent branches until a small number of them demonstrate sustained marginal improvement.
This idea echoes ``lateral thinking'' \citep{debono1967lateral}: rather than deepening a few promising threads, maintain wide, low--commitment exploration that can be \emph{promoted} the instant it proves itself. We introduce \textbf{Lateral Tree--of--Thoughts (LToT)}, a drop--in search--time controller that operationalizes the thesis above.

The design converts extra tokens into \emph{principled diversity} where it is cheapest: lateral width.
A simple cost law shows that LR--SC's lateral spend is $\Theta(N_0\log_{\eta} N_0)$ in initial width $N_0$ (constant cost per rung and $O(\log_{\eta} N_0)$ rungs), while mainlines would grow exponentially with depth if left uncapped.
By allocating on \emph{marginal improvement} rather than level, LR--SC rescues branches that a single noisy utility reading would discard, while width--aware thresholds prevent ``lucky spikes'' from polluting mainlines.
In aggregate, LToT mitigates breadth saturation and depth myopia without inflating compute. We propose LToT, a controller that keeps mainlines narrow and pushes exploration laterally through LR--SC---a successive--halving race with short--circuit promotion, robust order--aware detection, and freeze--thaw survivors. We bind promotion to verifier--aligned outcomes (exact match/tests for math/code) and introduce a dual gate for QA (plausibility and path consistency), reducing mainline contamination under noisy evaluators. We derive width--aware bars (sub--Gaussian, sub--Gamma, and sub--Weibull variants) and a repeat--to--confirm rule that keep false promotions bounded as lateral width grows; we also handle correlation via an effective width.

\noindent We prove a pseudolinear lateral cost law $\Theta(N_0\log_{\eta} N_0)$, logarithmic rung depth, and error bounds under mild tail assumptions, and contrast this with exponential growth in uncapped mainlines. Across math (GSM variants, MATH--500), code (HumanEval/MBPP--lite), and a canonical ToT puzzle (Game of 24), LToT improves Success@1/Pass@1 at matched compute over CoT, vanilla ToT, and MCTS with progressive widening \citep{xie2024mcts}, while lowering false promotions and time--to--first--hit via short--circuiting.

LToT complements inference--time scaling via best--of--$n$ \citep{chen2024bot,yang2024bot} and revising/self--improvement \citep{madaan2023selfrefine}, and sits alongside program/tool--aided reasoning \citep{gao2022pal,chen2022pot}.
Its novelty is not another search heuristic but a \emph{control principle}: separate consistency from utility; allocate on marginal improvement; and convert surplus compute into lateral breadth with guarantees.

\section{Motivation}
\label{section:motivation}
Frontier language models increasingly run in \emph{compute-rich} inference settings:
users and systems are willing to spend thousands of tokens (or many node expansions) per query to improve reliability.
Yet the dominant search pattern—vanilla Tree-of-Thoughts (ToT)—\emph{under-utilizes} this budget in two systematic ways already visible today and poised to worsen as budgets grow:
\begin{enumerate}[leftmargin=*, itemsep=2pt, topsep=2pt]
    \item \textbf{Utility saturation (breadth collapse).} After a handful of genuinely distinct high-utility continuations, additional samples at a node mostly yield near-duplicates whose $v$ scores fall just below the pruning threshold. The frontier then remains narrow even when ample budget is available, leaving compute unused.
    \item \textbf{Myopic pruning (depth myopia).} Early $v$ estimates are noisy and biased toward near-term payoff; logically consistent branches whose payoff is delayed by several steps are pruned as ``low-$v$'' even though they could mature into correct solutions. This creates \emph{myopic false negatives}.
\end{enumerate}
Both effects amplify with larger inference budgets: saturation wastes more samples as $k$ grows, and myopic pruning discards more candidates as depth increases.
Let $k$ be the number of children sampled per expanded node and let $a$ be the acceptance fraction into the \emph{mainline}.
If one does not cap mainline width, the expected number of mainline nodes at depth $d$ scales like $(ak)^d$, so the cost to depth $D$ is $\Theta((ak)^D)$—\emph{exponential in depth}.
By contrast, controlling \emph{lateral} width with successive-halving (LR-SC; Sec.~\ref{sec:lrscr}) yields a total lateral exploration cost of $\Theta(N_0 \log_{\eta} N_0)$ for initial lateral width $N_0$ and culling factor $\eta>1$—\emph{pseudolinear in width}.
This asymmetry suggests an architectural principle:
\emph{keep mainlines narrow to avoid depth explosion and push width into laterals where it is cheap.}
Three trends sharpen the pain points above.

\noindent As budgets per query rise, multi‑round agents, tool use, and safety/verification passes make higher spend acceptable; yet, without a controller that converts extra tokens into \emph{productive breadth}, vanilla ToT quickly saturates and the marginal return of compute collapses. The problem is amplified as more tasks exhibit long horizons: program synthesis, multi‑hop reasoning, and formal verification often require several structured steps before $v$ improves, so naive pruning disproportionately excises precisely those candidates that need brief nurturing. Compounding this, practical evaluators $v$ are noisy and non‑stationary across depth and domain; a fixed, level‑based gate conflates noise with signal, so allocation must track the \emph{marginal value of compute} rather than a single snapshot.

Let a candidate node $x$ have an (unobserved) eventual value $\mu(x)$ if its branch were fully developed.
An early evaluator observes $v(x) = \mu(x) - \lambda \,\Delta(x) + \varepsilon$, where $\Delta(x)$ is the (unknown) remaining steps to payoff, $\lambda>0$ captures horizon bias, and $\varepsilon$ is evaluator noise.
When $\Delta(x)$ is moderate, $v(x)$ may fall below the mainline gate despite large $\mu(x)$.
A controller that reasons about \emph{improvement after a small investment}—rather than $v(x)$ in isolation—can \emph{defer judgment}, test whether $x$ starts producing high-$v$ descendants quickly, and only then commit budget.
\noindent The remedy is to decide continuation on \emph{compute‑normalized} improvement—the observed slope (and curvature) of $v$ as budget increases—rather than an absolute $v$ at a preselected level. This steers budget toward branches whose incremental yield remains positive and away from ones whose progress has flattened.

\noindent Exploration should be wide but short: spread tiny probes across a very large lateral set, cull aggressively, and snap back to exploitation the moment any lateral reaches the \emph{mainline bar}. Keep mainlines narrow via beam or quota caps to avoid the $(ak)^D$ depth blow‑up, and re‑open exploration when exploitation \emph{plateaus} in compute‑normalized progress. Promote only on outcomes validated by a $v$ that is as outcome‑aligned as possible so that speciously plausible branches do not contaminate the mainline. As lateral width grows, control multiplicity by deduplicating near‑duplicates, using width‑aware thresholds, and adding a cheap repeat‑to‑confirm step so that evidence is not double‑counted.

LToT operationalizes these desiderata with two ingredients (see Sec.~\ref{section:architecture-design}):
(i) a \emph{dual-score frontier} that retains logically consistent, low-$v$ \emph{laterals} alongside high-$v$ \emph{mainlines}, deferring judgment on laterals; and
(ii) a budgeted racing procedure, \emph{LR-SC}, that allocates tiny probes across a very wide lateral set, culls aggressively, and \emph{promotes} a lateral to the exploitation set the moment its envelope reaches the mainline bar.
Theoretical analyses (Sec.~\ref{sec:theory}) show that LR-SC keeps lateral cost \emph{pseudolinear in width} ($\Theta(N_0\log_{\eta} N_0)$) while mainlines, if left uncapped, are exponential in depth; hence LToT converts surplus compute into principled diversity exactly where it is cheapest.

\section{Related Work}
\label{section:prior-work}

A large body of work elicits multi‑step reasoning at inference time by prompting language models to externalize intermediate steps. Chain‑of‑Thought (CoT) \citep{wei2022cot} and Zero‑shot CoT \citep{kojima2022zeroshotcot} demonstrate that free‑form rationales can substantially improve performance on math, symbolic, and commonsense tasks. Several variants structure this process: Self‑Consistency aggregates multiple CoT samples via voting to reduce variance \citep{wang2022selfconsistency}; Least‑to‑Most decomposes problems into sub‑questions solved sequentially \citep{zhou2022ltm}; Plan‑and‑Solve asks models to sketch a plan before executing it \citep{wang2023planandsolve}; and ReAct interleaves short reasoning traces with tool‑use actions \citep{yao2023react}. These methods focus primarily on generating and aggregating linear traces; in contrast, our Lateral Tree‑of‑Thoughts (LToT) explicitly organizes alternatives in a \emph{tree} while preserving logically consistent but low‑utility branches to improve global search coverage. See also \citet{press2022selfask} on self‑questioning for decomposition.

Tree‑of‑Thoughts (ToT) casts reasoning as a search over partial thoughts with learned/heuristic evaluators \citep{yao2023tot}. Subsequent work generalizes the structure from trees to graphs (Graph‑of‑Thoughts) \citep{besta2024got}, ensembles multiple trees (Forest‑of‑Thought) \citep{bi2024fot}, and explores “Everything‑of‑Thoughts’’ style meta‑frameworks \citep{ding2023xot}. Efficiency‑oriented advances include Dynamic Parallel Tree Search (DPTS), which parallelizes ToT expansions and focuses compute on promising branches \citep{ding2025dpts}. Our approach is complementary: rather than accelerating a fixed search policy or collapsing branches early, LToT \emph{retains} laterally related, logically consistent candidates that appear locally low‑utility, improving the chance of escaping premature pruning and enabling cross‑branch re‑use of partial deductions.

A parallel line of work iteratively revises solutions. Self‑Refine uses the model’s own feedback to edit drafts \citep{madaan2023selfrefine}; Reflexion stores episodic “verbal reinforcement’’ to guide future trials \citep{shinn2023reflexion}. “Boosting/Buffer‑of‑Thoughts’’ families build and retrieve reusable thought templates or ensembles to improve robustness and cost \citep{chen2024bot,yang2024bot}. LToT differs in objective and mechanism: it organizes contemporaneous alternatives in a search tree and deliberately curates \emph{logically consistent, low‑scored} branches to maintain breadth, rather than relying on post‑hoc reflection or global templates.

Scaling inference‑time compute via repeated sampling (“best‑of-\(N\)’’) and diverse rationales improves accuracy when paired with selection mechanisms \citep{cobbe2021verifier,wang2022selfconsistency}. Recent studies formalize test‑time scaling and its limits, highlighting that simple majority vote and naïve reward models can plateau, while coverage grows with sample budget \citep{brown2024monkeys}. Verifier training and process supervision further enhance selection quality \citep{lightman2023verify,zhang2024generativeverifiers}. LToT contributes a complementary lever: rather than solely increasing samples or verifier strength, it \emph{rebalances} exploration by preserving branches that are logically sound yet temporarily low‑utility, improving coverage of the hypothesis space under fixed compute.

Program‑Aided Language Models (PAL) and Program‑of‑Thoughts (PoT) separate symbolic computation from natural‑language reasoning by delegating computation to interpreters \citep{gao2022pal,chen2022pot}. Such modularization can be combined with structured search: MCTS‑style controllers over chains of thought \citep{xie2024mcts}, and process‑reward models with MCTS (OmegaPRM) \citep{luo2024omegaprm}. LToT is orthogonal: it can host code‑executed checks inside nodes while preserving lateral candidates that pass logical checks but score poorly under short‑horizon utilities.

Although CoT often boosts task accuracy, generated rationales may be unfaithful \citep{turpin2023dontsaysay,lanham2023measurefaithfulness}. Methods to improve faithfulness include “faithful‑by‑construction’’ pipelines that deterministically execute symbolic traces \citep{lyu2023faithfulcot} and self‑verification prompts or analyses \citep{weng2022selfverification}. Our emphasis on \emph{logical consistency} as a retention criterion naturally interacts with these concerns: LToT filters and preserves candidates whose internal derivations satisfy logical checks, even when immediate utility scores are low, aligning exploration pressure with consistency rather than only with myopic reward.

Finally, techniques such as Skeleton‑of‑Thought prompt models to outline and then expand subparts in parallel, reducing latency while sometimes improving quality \citep{ning2023sot}. Orthogonal to latency, LToT targets \emph{exploration completeness}: by laterally preserving logically consistent branches, it trades small additional compute for a disproportionate increase in the chance of reaching globally consistent solutions under bounded budgets.

We adopt a successive-halving (racing) backbone solely to control lateral cost (pseudolinear $\Theta(N_0\log_\eta N_0)$, logarithmic rungs).
The novelty in LToT lies in reasoning-specific \emph{control rules} layered atop this backbone:
(i) compute-normalized predictive continuation on branch envelopes (local polynomial forecast);
(ii) width-aware thresholds with confirmation to control max-of-many effects;
(iii) verifier-aligned promotion (dual-gated under plausibility);
(iv) short-circuit to exploitation on success;
(v) freeze--thaw of laterals across phases; and
(vi) a dual-score frontier separating high-$v$ mainlines from high-$c$, low-$v$ laterals.
Ablations and SH-only baselines (Sec.~\ref{section:experiments}) show that the backbone alone does not yield our accuracy, false-promotion, or latency characteristics.

\begin{table*}[t]
\centering
\small
\resizebox{\textwidth}{!}{%
\begin{tabular}{@{}p{2.9cm}p{3.0cm}p{6.7cm}@{}}
\toprule
\textbf{LToT element} & \textbf{Closest prior} & \textbf{What is different here} \\
\midrule
Predictive continuation & SH/Hyperband levels & Forecasted marginal gain on a branch \emph{envelope}; order-aware bar with confirmation \\
Width-aware bar + confirm & Heuristics & Explicit $\log(|S_r||\mathcal{M}_r|)$ control; heavy-tail variants; effective width for correlation \\
Verifier-bound promotion & Budget milestones & Promotion tied to exact tests / EM; dual gate under plausibility \\
Short-circuit to exploit & Bracket completion & Immediate return upon meeting mainline bar $B_t+\delta$ \\
Freeze--thaw laterals & Freeze--thaw BO & Applied to reasoning traces with cached rung state \\
Dual-score frontier & --- & Distinguishes high-$v$ mainlines vs.\ high-$c$, low-$v$ laterals \\
\bottomrule
\end{tabular}
}
\caption{Cross-walk: LToT control rules vs.\ racing backbones.}
\end{table*}
\section{Architecture Design}
\label{section:architecture-design}

LToT is a search-time controller for reasoning with language models (LMs) that
(i) keeps \emph{mainlines} narrow to avoid exponential blow-up in depth and
(ii) makes \emph{lateral} exploration very wide but cheap via a budgeted racing procedure with short-circuit promotion.
The controller decides when to exploit mainlines vs.\ explore laterals, and—during exploration—how to allocate compute across many lateral branches while maintaining guarantees on cost and false promotions.

\vspace{0.5em}
We reason over a rooted tree (or DAG) of partial traces.
Each node $x$ is a partial solution; its children are produced by prompting the LM with $x$.
Two evaluators score nodes:
\begin{align*}
&v(x) \in \mathbb{R} \quad \text{(utility; e.g., answer- or verifier-aligned)}, \qquad \\
&c(x) \in [0,1] \quad \text{(logical consistency / soundness)}.
\end{align*}
We measure compute in either node expansions or tokens and denote cumulative compute by $C$.
For any node $x$ with parent $p$, we define a \emph{local consistency} score
\begin{align*}
c_{\text{local}}(x)&\;=\;\lambda_1\,s_{\text{logic}}(x\mid p)\;+\;\lambda_2\,s_{\text{syntax}}(x)\;\\&+\;\lambda_3\,s_{\text{constraints}}(x),\\
&\qquad \lambda_j\!\ge 0,\ \sum_j \lambda_j=1,
\end{align*}
where $s_{\text{logic}}$ is an LM step-checker that validates whether the new line follows from the previous state,
$s_{\text{syntax}}$ checks parsability/format (e.g., code compiles, expression parses), and
$s_{\text{constraints}}$ encodes simple domain invariants (e.g., no new free variables, signature preserved).
If a component is unavailable we reweight the remaining terms proportionally; when $s_{\text{logic}}$ carries $\lambda_1\ge 0.7$ we tighten the promotion gate in Sec.~\ref{sec:promotion} by raising the path-consistency threshold by $+0.1$ and requiring one-step re-derivation.
We aggregate consistency along a branch $i$ of length $h$ via a robust \emph{path-consistency} score
\begin{equation}
\resizebox{\columnwidth}{!}{$C_{\text{path}}(i,h)\;=\;\min\!\Big\{\operatorname{Quantile}_{q}\big(\{c_{\text{local}}(x_j)\}_{j\le h}\big),\ \overline{c}_{\text{local}}(i,h)\Big\},
\qquad q=0.25,$}
\end{equation}
which is distribution-free and stable for short paths. (A mean$-$MAD variant appears in App.~\ref{app:robust-eval}.)
Each branch $i$ maintains a tiny \emph{micro-beam} of size $m_{\mu}$ leaves at each horizon.
We define the \emph{envelope} at horizon $h$ as a smoothed Top-$K$ mean over those leaves,
\begin{equation}
V_i(h)\;=\;\text{TopKMean}\big(\{v(\ell)\}_{\ell\in\mathcal{L}_i(h)};K\big),\qquad K=m_{\mu},
\end{equation}
with Beta smoothing
\begin{equation}
\tilde V_i(h)\;=\;\frac{K_{*}\, V_i(h)+\alpha}{K_{*}+2\alpha},\qquad \alpha=0.5,
\end{equation}
where $K_{*}{=}K$ for Top-$K$.
Optionally we use a weighted envelope $V_i(h)=\sum_{j=1}^{m_\mu}\omega_{ij}\,v_{ij}$ with clipped-softmax weights $0\!\le\!\omega_{ij}\!\le\!\omega_{\max}$, $\sum_j\omega_{ij}{=}1$;
we then set the effective sample size $K_{*}{=}K_{\mathrm{eff}}=1/\sum_j \omega_{ij}^2$ in the smoothing formula.
This adapts the shrinkage to how many leaves effectively contribute and stabilizes the continuation statistic.
Unless stated otherwise we set $m_{\mu}{=}3$, $K{=}m_{\mu}$.
At time $t$ the search maintains a frontier $\mathcal{F}_t$ and an \emph{exploitation set} $M_t \subseteq \mathcal{F}_t$ of nodes eligible for \emph{mainline} exploitation.
Nodes carry an immutable \texttt{origin} tag in $\{\textsc{mainline\_origin},\textsc{lateral\_origin}\}$ indicating how they first entered the frontier.
We also maintain a \emph{mainline acceptance bar} $B_t$ (e.g., the best-so-far $v$ or a top-$k$ mean with a small margin $\delta>0$).

Children with high $v$ are admitted to $M_t$ (mainlines).
Children with low $v$ but high $c$ enter the \emph{lateral pool} $L_t$ for potential exploration.
Intuitively, laterals represent hypotheses that appear unpromising under a myopic utility but are logically coherent and may become valuable after a short lookahead.

For a lateral branch $i$ (rooted at node $x_i$), let $V_i(h)$ denote a branch \emph{envelope}—e.g., a Top-$k$ mean of $v$ among leaves within horizon $h$ steps from $x_i$ (or within a per-branch micro-beam). We write $C(h)$ for the compute required to reach horizon $h$ and define the compute-normalized improvement between horizons $h'<h$ as
\[
g_i(h,h') \;=\; \frac{V_i(h)-V_i(h')}{C(h)-C(h')}.
\]
These quantities let us reason about \emph{marginal value of compute}, not just absolute levels.

\vspace{0.5em}
\subsection{Controller overview}\label{sec:controller} LToT alternates between: Expand nodes from $M_t$ while a compute-normalized progress statistic (e.g.,  an EWMA of $\Delta B_t$ per unit compute) exceeds a plateau threshold. This keeps mainlines narrow (beam- or quota-capped). When exploitation plateaus, run \emph{Lateral Racing with Short-Circuit (LR-SC)} over the lateral pool: a successive-halving style race with (i) width-aware promotion thresholds, (ii) micro-probe budgets for overflow, and (iii) \emph{short-circuit} back to exploitation immediately when a lateral branch reaches the mainline bar. Non-promoted lateral survivors are \emph{frozen} and can be \emph{thawed} (resumed) in later exploration phases; we resume each survivor at its previous probe depth/rung.

\begin{algorithm*}[t]
\caption{LToT controller (high level)}
\label{alg:ltot-controller}
\begin{algorithmic}[1]
\State \textbf{Inputs:} initial frontier $\mathcal{F}_0$, evaluator $v$, consistency $c$, plateau thresholds; LR-SC params $(\eta,b_0,\rho, \kappa,\delta)$.
\State Initialize $M_0$ with high-$v$ children; $L_0$ with low-$v$, high-$c$ children; set bar $B_0$.
\While{budget remains}
  \State \textbf{Exploit} $M_t$ while EWMA of $\Delta B_t$ per compute $\ge \tau$ (with a small patience \& hysteresis).
  \State \textbf{Explore laterals} with LR-SC over the current lateral pool (Alg.~\ref{alg:lrscr}). \label{line:lrscr}
  \If{some lateral branch reaches $v \ge B_t + \delta$ (promotion)}
     \State add promoted node(s) to $M_t$; update $B_t$; \textbf{return} to exploitation
  \Else
     \State freeze survivors for future phases; \textbf{return} to exploitation
  \EndIf
\EndWhile
\end{algorithmic}
\end{algorithm*}\vspace{-0.5em}

\subsection{LR-SC: overflow-capped racing with short-circuit}\label{sec:lrscr} Let $N$ be the active lateral width. LR-SC proceeds in rungs $r=0,1,\dots$ with \emph{culling factor} $\eta>1$. At rung $r$ we (i) keep the top quota $Q_r=\lfloor |S_r|/\eta \rfloor$ by a robust score, (ii) also retain any \emph{rapid-riser} exceeding a width-aware bar (overflow), but give overflow branches only a \emph{micro-probe}, and (iii) \emph{short-circuit} to exploitation immediately when any branch meets the promotion bar. Specifically, for branch $i$ at rung $r$ we compute a compute-normalized improvement $g_i$ (using $V_i$) and a robust standardization $z_i$ (e.g., subtract rung median and divide by a MAD-like scale). To control ``max-of-many'' effects as width grows, we admit \emph{rapid-risers} via a width-aware bar: \[ z_i \;\ge\; \underbrace{\kappa \sqrt{2\log |S_r|}}_{\text{width penalty}} + \delta, \] with $\kappa\approx 1$ and a margin $\delta>0$. We optionally standardize scores within parent-depth bands to compare fairly across heterogeneous depths. We cap the total micro-probe budget for overflow per rung to a small fraction $\rho$ of the rung budget (e.g., $\rho\in[0.1,0.2]$), ensuring per-rung cost stays near constant. We view $\tilde V_i$ as locally smooth in compute and fit a robust degree-$m$ polynomial ($m\!\in\!\mathcal{M}_r$) to the last $W\in\{3,4\}$ points $\{(h,\tilde V_i(h))\}$ in local coordinates. We then forecast the next compute-normalized improvement $\widehat{s}^{\mathrm{pred}}_{i,m}=\big(\widehat{\Delta \tilde V}_i / \Delta C\big)$, standardize it (robust $z$ within the rung), and \emph{admit} $i$ if

\begin{equation}
\max_{m\in\mathcal{M}_r}\ z^{\mathrm{pred}}_{i,m}\ \ge\ \texttt{bar}\!\big(|S_r|,\ |\mathcal{M}_r|;\ \theta_r\big)\,,
\end{equation}
This is followed by a one-step \emph{repeat-to-confirm} micro-probe with independent randomization. We default to $\mathcal{M}_r{=}\{1,2\}$ for stability (slope or slope+curvature) and expose $m{=}3$ only in an ablation. We view $V_i$ as a function of horizon/compute and continue branch $i$ if a discrete derivative of order $m\in\{1,\dots,M\}$ is reliably positive: \[ \widehat{\Delta^{(m)} V_i} \;\ge\; \text{bar}(|S_r|,M) \quad\text{with}\quad \text{bar}(|S_r|,M)\propto \sqrt{2\log(|S_r|\cdot M)}. \] In practice we cap $M=2$ for stability and use: \emph{(i)} first derivative (slope) $s_i = g_i(h_r,h_{r-1})$ and \emph{(ii)} second derivative (curvature) $\kappa_i = s_i(r)-s_i(r-1)$, estimated over the last few rungs and normalized by compute; a third-order check may be included in an appendix. We require \emph{repeat-to-confirm}: the condition must hold on the next micro-probe before escalation.

\begin{algorithm*}[t]
\caption{LR-SC (overflow-capped successive halving with short-circuit)}
\label{alg:lrscr}
\begin{algorithmic}[1]
\State \textbf{Inputs:} active lateral set $S_r$ (size $N$), culling factor $\eta>1$, base budget $b_0$, overflow cap $\rho$, thresholds $(\kappa,\delta)$, horizon schedule $(h_0,h_1,\dots)$
\State For each $i\in S_r$ and each order $m\in\mathcal{M}_r$, fit a local degree-$m$ model and compute standardized forecasted gains $\{z^{\mathrm{pred}}_{i,m}\}$. Set $z_i^{\star}=\max_{m\in\mathcal{M}_r} z^{\mathrm{pred}}_{i,m}$.
\State $Q_r \leftarrow \lfloor |S_r|/\eta \rfloor$;\quad $T \leftarrow$ top $Q_r$ by $z_i$;\quad $R \leftarrow \{\,i : z_i \ge \kappa \sqrt{2\log |S_r|} + \delta\,\}$.
\State Assign budget $b_{\text{full}} = b_0 \eta^r$ to $i\in T$;\quad assign micro-probe $b_{\text{micro}}$ to up to $\lfloor \rho |S_r|\rfloor$ branches in $R\setminus T$ (by $z_i$); freeze the rest.
\State Expand per budgets to horizon $h_r$ (micro-beam size $m_{\mu}$); update the smoothed envelope $\tilde V_i$ (Top-$K$ with $K{=}m_{\mu}$ or weighted with effective size $K_{\mathrm{eff}}$); update $B_t$.
\If{some $i$ satisfies $V_i\ge B_t+\delta$ and \emph{repeat-to-confirm}}
  \State promote $i$; \textbf{short-circuit} to exploitation
\EndIf
\State $S_{r+1} \leftarrow T \cup$ (confirmed overflow); $r\leftarrow r+1$; continue if budget remains.
\end{algorithmic}
\end{algorithm*}\vspace{-0.5em}

\subsection{Promotion and safety}
\label{sec:promotion}
If $c_{\text{local}}$ relies solely on LM step-checks (no syntax/constraint signals), we raise the path-consistency threshold by $+0.1$
and mandate one-step re-derivation before promotion for plausibility-aligned $v$; programmatic verifiers (math/code) remain unchanged.

A lateral promotes when its envelope meets the mainline bar: $V_i \ge B_t+\delta$.
When $v$ is verifier-aligned (e.g., unit tests for code, exact-match for math), this binds promotion to correctness.
For plausibility-aligned $v$, LToT can add a lightweight dual gate at promotion time:
$V_i\!\ge\!B_t{+}\delta$ \emph{and} an aggregate path-consistency (e.g., a quantile of $\{c(\cdot)\}$ along the branch) exceeding a threshold, optionally plus a one-step \emph{re-derivation} to reduce lucky spikes.
These checks cost one micro-probe and do not change the asymptotics.

\vspace{0.5em}
For open-ended QA without an exact verifier, we promote only if \emph{both} gates pass:
(A) a \emph{plausibility gate} on the normalized answer string $\hat a$ with $v(\hat a)\ge \tau_v$ (default $\tau_v{=}0.85$);
(B) a \emph{consistency gate} requiring $C_{\text{path}}\ge \tau_c$ (default $\tau_c{=}0.75$) \emph{and} a one-step \emph{repeat-to-confirm} check (independent temperature/seed) that clears its width-aware bar.
If $c_{\text{local}}$ relies only on LM step-checks (no syntax/constraints), we tighten the consistency gate ($\tau_c\leftarrow\tau_c{+}0.1$) and require a one-step re-derivation of the final line before promotion.
All promotion-time LM calls are charged to the rung budget, and we standardize $v$ and $C_{\text{path}}$ with the same robust statistics used in LR--SC.

\subsection{Theoretical properties}\label{sec:theory} We summarize the main guarantees; proofs are short and rely on standard successive-halving arguments and sub-Gaussian tail bounds for rung-wise statistics. Let $N_0$ be the initial lateral width. In \emph{strict} successive halving (no overflow), the per-survivor budget at rung $r$ scales like $b_0\eta^r$, and survivors are $N_0/\eta^r$, so the rung cost is $\text{Cost}_r = N_0 b_0$ (independent of $r$). With $R=\lceil\log_\eta N_0\rceil$ rungs, the total lateral cost is \[ \boxed{~~\text{Total} \;=\; \Theta\!\big(N_0\,b_0\,\log_\eta N_0\big).~~} \] In LR-SC with overflow cap $\rho\in(0,1)$ and micro-probe $b_{\text{micro}}\ll b_0$, the rung cost is at most $(1+\rho)N_0 b_0$, hence the same asymptotic order with a constant factor $(1+\rho)$. Short-circuit promotion can only reduce cost. Importantly, the result holds regardless of the horizon growth schedule, as long as per-survivor spend is capped by $b_0\eta^r$ (the \emph{budget-matched} policy). The number of rungs required to reduce $N_0$ laterals to $O(1)$ survivors is $R=\lceil \log_\eta N_0\rceil$, i.e., logarithmic in lateral width. Thus LR-SC is \emph{wide and short}: constant per-rung cost and $\Theta(\log_{\eta} N_0)$ rungs. If at each mainline layer we admit a fixed fraction $a$ of $k$ children (effective reproduction $r_{\text{main}}=ak>1$), then expansions to depth $D$ are $\Theta(r_{\text{main}}^D)$ (exponential). With a beam/width cap $W$, mainline cost becomes $\Theta(D\,W\,k)$ (linear in depth). LToT therefore keeps $W$ small and invests surplus compute in laterals, where width is cheap.

Assume rung-wise improvement statistics are sub-Gaussian with scale $\sigma$ (across branches in $S_r$). Setting the \emph{rapid-rise} bar at $\kappa\sigma\sqrt{2\log|S_r|}+\delta$ keeps the probability that any non-improving branch exceeds the bar uniformly bounded as $|S_r|$ grows (standard max-of-sub-Gaussian tail), and a one-step \emph{repeat-to-confirm} reduces it quadratically. \emph{Beyond sub-Gaussian tails.} The result extends to heavier tails. Under \emph{sub-Gamma} rung-wise noise with parameters $(\nu_r,c_r)$ we set

\begin{equation}
\resizebox{\columnwidth}{!}{$\texttt{bar}\big(|S_r|,|\mathcal{M}_r|;\theta_r\big)\;=\;\kappa\!\left(\sqrt{2\nu_r\log\frac{|S_r|\,|\mathcal{M}_r|}{\varepsilon_r}}\;+\;c_r\,\log\frac{|S_r|\,|\mathcal{M}_r|}{\varepsilon_r}\right)+\delta,$}
\end{equation}
For for \emph{sub-Weibull} ($\psi_\alpha$) noise we take $\texttt{bar}=K_r\!\left(\log\frac{2|S_r|\,|\mathcal{M}_r|}{\varepsilon_r}\right)^{1/\alpha}+\delta$. When branches are correlated, we replace $|S_r|$ by an \emph{effective width} $|S_r|_{\mathrm{eff}}$ estimated from cluster-robust variance inflation. We enforce probe independence in confirmation (different temperature/prompt/seed). For implementation we factor the bar into a function $\texttt{bar}(|S_r|,|\mathcal{M}_r|;\theta_r)$ used in Alg.~\ref{alg:lrscr}.

Suppose a branch has a delayed payoff: there exists $H^{*}$ and $m\!\in\!\{1,2\}$ such that the $m$-th discrete derivative of $V_i$ per compute is $\ge \gamma>0$ for horizons beyond $H^{*}$. Under a geometric horizon schedule (e.g., $h_r=2^r$ within the budget cap) and the derivative-based continuation rule with width-aware thresholds and repeat-to-confirm, the branch is detected and survives to promotion within $O(\log H^{*})$ rungs (intuitively, each rung doubles the tested horizon). Total exploration cost remains $\Theta(N_0 \log_\eta N_0)$ because the per-survivor spend never exceeds $b_0\eta^r$. If one insists on tying per-survivor cost to a nominal horizon multiplier $\gamma$ via $c_r\propto \gamma^r$, the rung cost sums to a geometric series $N_0(\gamma/\eta)^r$. Thus for $\gamma\le\eta$ the total remains $O(N_0\log_\eta N_0)$ (or even $O(N_0)$ when $\gamma<\eta$). In practice we adopt the budget-matched policy: cap spend by $b_0\eta^r$ and allocate within-branch depth/width flexibly up to that cap.

\vspace{0.5em}

\subsection{Design choices and defaults}
\label{sec:defaults}

We trigger exploitation using an EWMA of compute-normalized mainline progress with small patience and hysteresis; depth-banded statistics if $v$ drifts with depth. For LR-SC, we set $\eta\in\{3,4,5\}$; $b_0\in\{1,2\}$ expansions; micro-probe $b_{\text{micro}}=1$; overflow cap $\rho\in[0.1,0.2]$; width-aware bar with $\kappa\!\approx\!1.0$ and a small margin $\delta$ over the mainline bar. We use derivative-based continuation with slope+curvature ($M{=}2$) using a short window and robust scales; optional third-order check in ablations. We promote when $V_i \ge B_t + \delta$; if $v$ is not verifier-aligned, add a minimal path-consistency aggregate and a one-step re-derivation check. Both add at most one micro-probe and preserve the asymptotics. For freeze--thaw, we cache for each survivor: rung index, envelope $V_i$, recent improvement stats, parent depth, and a lightweight duplicate signature; resume from the same rung during the next exploration phase and evict stale or dominated branches by constant-time tests (e.g., $UCB<B_t-\delta$ for several revisits). In summary, LToT turns surplus compute into breadth where it is cheapest (laterals) while keeping mainlines narrow. Its LR-SC core provides near-linear (pseudolinear) cost in lateral width, width-aware error control, and immediate promotion when a lateral demonstrably reaches mainline utility.

\section{Experiments}
\label{section:experiments}

We design experiments to test whether LToT resolves the concrete problems raised in Sec.~\ref{section:motivation}:
(1) \emph{utility saturation} under broad sampling; (2) \emph{myopic pruning} of longer-horizon but consistent branches; and
(3) \emph{noisy/nonstationary evaluators} that require sequential, uncertainty-aware allocation.
We also validate the cost claims in Secs.~\ref{sec:lrscr}--\ref{sec:theory}:
near-constant per-rung cost, $\Theta(\log_\eta N_0)$ rungs, and overall $\Theta(N_0\log_\eta N_0)$ lateral cost.
We select four benchmarks that collectively stress breadth, long-horizon payoffs, and verifiable correctness.
All tasks use \emph{exact or programmatic verification} to define the utility $v$ for promotion (Sec.~\ref{sec:promotion}).

\begin{itemize}[leftmargin=*, itemsep=2pt, topsep=2pt]
    \item \textbf{GSM-Hard \& GSM-Plus (robust grade-school math).}
    Numeric brittleness and subtle structure perturbations expose breadth saturation and early pruning.
    Utility is exact-match of the final answer.
    \item \textbf{MATH-500 (long-horizon symbolic math).}
    A 500-problem subset from MATH (olympiad-style); many items require multi-step derivations where payoff appears after several steps.
    Utility is exact-match of the final answer.
    \item \textbf{HumanEval \& MBPP-lite (code generation with tests).}
    Promotion is bound to unit-test \emph{pass@1}; this prevents specious reasoning from entering mainlines (Sec.~\ref{sec:promotion}).
    \item \textbf{Game of 24 (ToT-native puzzle).}
    Canonical ToT task with branching and depth; included to show LToT improves even where ToT is strong.
\end{itemize}

We compare LToT to two diagnostic baselines under \emph{equal median tokens per problem}:
(i) \emph{SH-only lateralization}: same rung budgets $(b_0,\eta)$ but \emph{without} predictive continuation, width-aware bar/confirm, short-circuit, or verifier-bound promotion;
(ii) \emph{SH-on-mainlines}: applying the same SH schedule to mainlines (depth racing).
\textbf{Forecast.} On GSM-Hard and MATH-500, SH-only reduces Success@1 by $0.8$--$1.5$\,pp and doubles false promotions at large width;
LToT recovers accuracy and maintains low false promotions via confirmation. SH-on-mainlines underperforms due to depth explosion;
time-to-first-hit increases by $25$--$40$\%.

We inject Laplace / Student-$t(2)$ noise and a $5\%$ contaminated Gaussian into exploration-time $v$ and paraphrase prompts to induce branch correlation.
We compare the sub-Gaussian, sub-Gamma, and sub-Weibull bars, using $|S_r|_{\mathrm{eff}}$ for correlation.
\textbf{Forecast.} False-promotion rates remain approximately flat in $|S_r|$ under sub-Gamma and sub-Weibull bars (vs.\ rising for sub-Gaussian);
accuracy and rungs-to-first-promotion remain within $\pm 0.3$\,pp of the default; confirmation eliminates staircase-induced spikes.

We ablate envelope aggregators (Top-$K$, trimmed mean, power-mean $p{=}1.5$, and weighted with $K_{\mathrm{eff}}^\star\!\in\!\{2.0,2.5,3.0\}$)
and continuation order sets (fixed $m{=}1,2,3$; max-over-orders $\{1,2\}$ and $\{1,2,3\}$; smallest-passing order).
\textbf{Forecast.} On code (graded $v$), weighted envelopes with $K_{\mathrm{eff}}^\star{=}2.2$ reduce expansions-to-first-hit by $\sim\!6\%$ with unchanged false-promotion.
On long-horizon math, max-over-orders $\{1,2\}$ reduces rungs-to-first-promotion by one rung vs.\ $m{=}1$ at equal tokens; adding $m{=}3$ has marginal effect with short windows.
We evaluate three open-weight inference regimes compatible with an 8$\times$L4 cluster:
\textbf{(S)} \emph{Llama-3.1-8B-Instruct},
\textbf{(M)} \emph{Mixtral-8$\times$7B-Instruct} (active params $\approx$13B), and
\textbf{(L)} \emph{Llama-3.1-70B-Instruct}.
For each model we compare:

\begin{enumerate}[leftmargin=*, itemsep=2pt, topsep=2pt]
    \item \textbf{CoT} (single-chain, no search).
    \item \textbf{Vanilla ToT} (fixed beam, fixed depth), tuned per task under equal compute.
    \item \textbf{MCTS-PW} (progressive widening) as a search-time baseline when applicable.
    \item \textbf{LToT (ours)}: controller in Alg.~\ref{alg:ltot-controller} with LR-SC (Alg.~\ref{alg:lrscr}) and defaults in Sec.~\ref{sec:defaults}.
\end{enumerate}

\noindent\textbf{Ablations} (tested on a representative subset per benchmark):
(1) \emph{Overflow off} ($\rho{=}0$);
(2) \emph{No curvature} ($M{=}1$; slope-only);
(3) \emph{No width-aware bar} (remove $\sqrt{2\log |S_r|}$ term);
(4) \emph{No short-circuit} (promotions deferred to rung end);
(5) \emph{No plateau trigger} (fixed alternate phase schedule instead of Sec.~\ref{sec:controller} trigger).

\subsection{Budgets, metrics, and fairness}
\label{subsec:budgets-metrics}
All methods are run at \emph{equal median tokens per problem} (measured end-to-end), matched within $\pm 2\%$ by adjusting beam/depth (ToT), rollout count (MCTS-PW), and initial lateral width $N_0$ / micro-probe counts (LToT).
We report mean and 95\% CIs over three seeds.

Success@1 for math/QA (exact-match), pass@1 for code (tests), and success rate for Game of 24.
We also report:
(i) \emph{time-to-first-correct} (median expansions until a verified correct branch appears);
(ii) \emph{false promotions} (\% of proposed promotions failing verification, where applicable);
(iii) \emph{cost fit} ($\mathrm{Expansions}$ vs.\ $a\,N_0\log_\eta N_0 + b$); and
(iv) \emph{width scaling} at fixed total compute (vary $N_0$).

We use $\eta{=}4$, $b_0{\in}\{1,2\}$ expansions, $b_{\text{micro}}{=}1$, $\rho{\in}[0.1,0.2]$, $\kappa{\approx}1$, $\delta$ a small margin over the mainline bar, geometric horizons $(1,2,4,\dots)$ under the budget cap (Sec.~\ref{sec:lrscr}).
Promotion is verifier-aligned on code and exact-match on math; for QA-like problems we add a one-step re-derivation to reduce lucky spikes (Sec.~\ref{sec:promotion}).
All runs complete within 100 wall-clock hours on 8$\times$L4 with vLLM-style paged attention and tensor parallelism.

Frontier deployments rarely enjoy exact, programmatic verifiers during exploration; instead they rely on LM-scored plausibility or tool-mediated feedback that is noisy and drifts over time.
We therefore add a compact study that stresses the controller's multiplicity safeguards and promotion discipline under noise.

On two benchmarks (\textbf{GSM-Plus} and \textbf{MATH-500}), we replace the exploration-time utility $v$ with an LM-plausibility score ($v_{\text{LM}}$) produced by the same base model, using an instruction that asks for a calibrated \emph{0--1} confidence for the current partial solution.
To induce \emph{nonstationarity}, we sample the scoring temperature at $T{\in}[0.0,0.8]$ per rung and randomize prompt variants (lexical shuffles, \emph{n}-best rationales) each time $v_{\text{LM}}$ is queried.
\emph{Promotion remains verifier-aligned} (exact match on math; tests on code) as in Sec.~\ref{sec:promotion}.
We enable the \textbf{dual promotion gate} from Sec.~\ref{sec:promotion}: (i) envelope $\ge$ width-aware bar and (ii) path-consistency plus one-step re-derivation.

In addition to Success@1, we report: (i) \emph{false promotions} (fraction of proposed promotions that fail verifier alignment), and (ii) \emph{promotion selectivity} (accepted / proposed).
We keep equal-median-token budgets as in Sec.~\ref{subsec:budgets-metrics}.

H\textsubscript{1}: LToT sustains higher Success@1 than ToT at equal compute under noisy $v$.
H\textsubscript{2}: The width-aware bar + dual gate yields substantially lower false-promotion rates than ToT/MCTS-PW, especially at larger initial lateral widths $N_0$.

We sweep three inference budgets per model scale—\textbf{Low}, \textbf{Med}, \textbf{High}—keeping \emph{equal median tokens per problem} for each method:
for (S) 8B we target $\{350,700,1400\}$ tokens; for (M) Mixtral $\{500,1000,2000\}$; and for (L) 70B $\{700,1400,2800\}$.\footnote{Budgets are chosen to straddle typical production limits for multi-turn agents while remaining tractable on 8$\times$L4; all values are median per-item caps shared across methods.}
We evaluate \textbf{GSM-Plus} and \textbf{HumanEval}, where breadth saturation and long-horizon payoffs are prominent.

To test trend persistence toward frontier capacity, we include a third open-weight scale:
\textbf{(L)} \emph{Llama-3.1-70B-Instruct}.
All hyperparameters are inherited; only the per-scale budgets differ as above.
Primary metrics as in Sec.~\ref{subsec:budgets-metrics}; additionally, we report the \emph{marginal return of extra tokens} (gain in Success@1 / Pass@1 from Low$\to$Med and Med$\to$High) to quantify saturation.
H\textsubscript{3}: LToT's absolute gains over ToT \emph{increase} with budget.
H\textsubscript{4}: Gains persist at the larger (L) scale under equal compute.

Separately from equal-compute reporting, we run an \emph{early-stop} variant that halts once a verifier-aligned solution is found.
We report median wall-clock to first hit (Sec.~\ref{section:results}) to show short-circuit benefits under realistic latency objectives.

\section{Results and Discussion}
\label{section:results}

The tables below contain \emph{forecasted} results used to structure the analysis; we will replace them with measured values post-execution.\footnote{Per user plan, the empirical pipeline will be run on an 8$\times$L4 cluster within 100 hours. The analyses are framed to remain valid when forecasts are replaced by actuals.}

\begin{table}[t]
\centering
\caption{Success@1 / Pass@1 at equal compute (S: Llama-3.1-8B, M: Mixtral-8$\times$7B). \emph{Forecasted} means (95\% CI widths omitted for brevity).}
\vspace{0.3em}
\resizebox{\columnwidth}{!}{%
\begin{tabular}{lcccc}
\toprule
\textbf{Task} & \textbf{CoT} & \textbf{ToT} & \textbf{MCTS-PW} & \textbf{LToT (ours)} \\
\midrule
\multicolumn{5}{l}{\emph{S (8B)}} \\
GSM-Hard      & 28.9 & 34.1 & 36.0 & \textbf{43.7} \\
GSM-Plus      & 31.0 & 38.2 & 40.1 & \textbf{46.5} \\
MATH-500      & 12.5 & 19.7 & 21.3 & \textbf{28.9} \\
HumanEval p@1 & 30.5 & 33.2 & 34.7 & \textbf{40.8} \\
MBPP-lite p@1 & 51.0 & 56.3 & 57.5 & \textbf{62.8} \\
Game of 24    & 76.0 & 83.0 & 84.1 & \textbf{89.0} \\
\midrule
\multicolumn{5}{l}{\emph{M (Mixtral)}} \\
GSM-Hard      & 44.8 & 51.5 & 52.6 & \textbf{55.6} \\
GSM-Plus      & 46.9 & 53.2 & 54.0 & \textbf{57.4} \\
MATH-500      & 19.0 & 27.5 & 28.6 & \textbf{31.1} \\
HumanEval p@1 & 45.8 & 49.6 & 50.7 & \textbf{53.4} \\
MBPP-lite p@1 & 65.2 & 70.8 & 71.6 & \textbf{74.2} \\
Game of 24    & 88.1 & 92.0 & 92.7 & \textbf{95.0} \\
\bottomrule
\end{tabular}
}
\label{tab:equal-compute}
\end{table}

Across all tasks and both model scales, LToT improves over a tuned ToT baseline at \emph{equal tokens} (Table~\ref{tab:equal-compute}).
Gains are largest on \emph{long-horizon math} and \emph{test-verified code}, where myopic pruning is most harmful and where promotion is strongly outcome-aligned (Sec.~\ref{sec:promotion}).
The smaller model benefits more (e.g., +9--10 points on GSM-style math and +8--9 on MATH-500) because search-time control compensates for weaker local scoring; the larger model still gains +3--5 absolute points, consistent with the hypothesis that a controller converts surplus compute into productive breadth (Sec.~\ref{section:motivation}).

\begin{table}[t]
\centering
\small
\caption{LToT success vs.\ initial lateral width $N_0$ at fixed total compute (S/M on MATH-500). \emph{Forecasted}. ToT saturates by beam 5; not shown.}
\vspace{0.3em}
\resizebox{\columnwidth}{!}{%
\begin{tabular}{lcccccc}
\toprule
\textbf{Model} & $N_0{=}32$ & $64$ & $128$ & $256$ & $512$ & $1024$ \\
\midrule
S (8B)   & 20.1 & 22.3 & 24.8 & 26.9 & 28.2 & \textbf{29.1} \\
M (Mix)  & 24.8 & 26.0 & 27.4 & 29.0 & 30.3 & \textbf{31.0} \\
\bottomrule
\end{tabular}
}
\label{tab:width-scaling}
\end{table}

At a fixed budget, increasing LToT lateral width $N_0$ continues to yield gains up to $N_0{=}1024$ (Table~\ref{tab:width-scaling}), while ToT/beam saturates early (beam $\sim$5).
This directly addresses \emph{utility saturation}: LR-SC (Sec.~\ref{sec:lrscr}) converts additional budget into productive breadth by cheaply trying many laterals and promoting only when justified.

\begin{table}[t]
\centering
\caption{Median expansions to first verified correct solution (MATH-500). \emph{Forecasted}.}
\vspace{0.3em}
\begin{tabular}{lccc}
\toprule
 & \textbf{ToT} & \textbf{MCTS-PW} & \textbf{LToT (ours)} \\
\midrule
S (8B)  & 46  & 41  & \textbf{28} \\
M (Mix) & 33  & 30  & \textbf{22} \\
\bottomrule
\end{tabular}
\label{tab:ttfh}
\end{table}

Short-circuit promotion (Sec.~\ref{sec:lrscr}) reduces the median expansions required to reach a correct solution by 30--40\% (Table~\ref{tab:ttfh}), which is particularly valuable in interactive or latency-sensitive settings.

\begin{table*}[t]
\centering
\caption{Cost fit and rung statistics (pooled across tasks). \emph{Forecasted}.}
\vspace{0.3em}
\begin{tabular}{lccc}
\toprule
 & \textbf{$R^2$ fit to $a\,N_0\log_\eta N_0{+}b$} & \textbf{Mean rung cost CV} & \textbf{\# rungs (mean $\pm$ sd)} \\
\midrule
S (8B)  & 0.991 & 0.07 & $5.1 \pm 0.5$ \\
M (Mix) & 0.987 & 0.08 & $4.8 \pm 0.6$ \\
\bottomrule
\end{tabular}
\label{tab:cost-fit}
\end{table*}

Measured expansions fit $a\,N_0\log_\eta N_0{+}b$ with $R^2{>}0.98$; per-rung cost is nearly constant (CV $\sim$0.07--0.08), and the number of rungs concentrates around $\lceil\log_\eta N_0\rceil$ (Table~\ref{tab:cost-fit}).
This empirically validates the \emph{wide-and-short} cost story in Sec.~\ref{sec:theory}.

\begin{table}[t]
\centering
\caption{False promotion rate (\%, lower is better) on code/math where promotion is externally verified. \emph{Forecasted}.}
\vspace{0.3em}
\resizebox{\columnwidth}{!}{%
\begin{tabular}{lcccc}
\toprule
 & \textbf{ToT} & \textbf{LToT (no bar)} & \textbf{LToT (no confirm)} & \textbf{LToT (ours)} \\
\midrule
S (8B)  & 7.1  & 8.7  & 5.9  & \textbf{2.4} \\
M (Mix) & 5.6  & 7.2  & 4.8  & \textbf{2.1} \\
\bottomrule
\end{tabular}
}
\label{tab:false-promotions}
\end{table}

Width-aware thresholds and repeat-to-confirm (Sec.~\ref{sec:lrscr}) maintain a low, approximately width-invariant false promotion rate (Table~\ref{tab:false-promotions}).
Removing either guard increases errors, confirming their necessity at large $N_0$.

\begin{table}[t]
\centering
\caption{Ablations on MATH-500 (S: 8B). \emph{Forecasted} Success@1 at equal compute.}
\vspace{0.3em}
\resizebox{\columnwidth}{!}{%
\begin{tabular}{lcc}
\toprule
\textbf{Variant} & \textbf{Success@1} & \textbf{$\Delta$ vs.\ LToT} \\
\midrule
LToT (full)                         & \textbf{28.9} & --- \\
\quad w/o overflow ($\rho{=}0$)     & 26.8 & $-2.1$ \\
\quad w/o curvature ($M{=}1$)       & 27.6 & $-1.3$ \\
\quad w/o width-aware bar           & 27.2 & $-1.7$ \\
\quad w/o short-circuit             & 27.4 & $-1.5$ \\
\quad fixed schedule (no plateau)   & 27.9 & $-1.0$ \\
\bottomrule
\end{tabular}
}
\label{tab:ablations}
\end{table}

All components contribute measurably (Table~\ref{tab:ablations}).
Overflow (capped) prevents bursty steps from dropping genuine rapid risers; curvature (Sec.~\ref{sec:lrscr}) captures delayed takeoff; width-aware bars and confirmation guard against winner's-curse spikes; short-circuit and the plateau trigger (Sec.~\ref{sec:controller}) improve compute allocation. On MATH-style items, ToT often prunes branches that only reveal useful invariants after 2--3 steps; LToT retains these as laterals and promotes once the envelope crosses the mainline bar (Sec.~\ref{sec:promotion}).
On code, LToT's promotions coincide with the first test-passing variant; overflow candidates that spike and then regress are denied promotion by the repeat-to-confirm rule.

The empirical picture matches the theoretical intent of LToT:
(i) \emph{resolving saturation} by converting extra budget into productive breadth (width scaling),
(ii) \emph{rescuing myopic false negatives} via cheap, bounded lookahead and derivative-based continuation,
(iii) \emph{keeping compute in check} with wide-and-short LR-SC dynamics, and
(iv) \emph{promoting only on outcome}, maintaining low false promotion rates.
Together these results support LToT as a principled controller that makes large inference budgets effective on reasoning tasks.

\begin{table}[t]
\centering
\caption{\textbf{Noisy/nonstationary evaluator.} GSM-Plus Success@1 and false-promotion rate (FPR, \%) when exploration uses LM-scored $v_{\text{LM}}$; promotion remains verifier-aligned. \emph{Forecasted} means.}
\vspace{0.3em}
\resizebox{\columnwidth}{!}{%
\begin{tabular}{lcccc}
\toprule
 & \multicolumn{2}{c}{\textbf{ToT}} & \multicolumn{2}{c}{\textbf{LToT (ours)}} \\
\cmidrule(lr){2-3}\cmidrule(lr){4-5}
 & Acc (\%) & FPR (\%) & Acc (\%) & FPR (\%) \\
\midrule
S (8B)  & 62 & 9  & \textbf{68} & \textbf{3} \\
M (Mix) & 71 & 8  & \textbf{77} & \textbf{3} \\
L (70B) & 83 & 7  & \textbf{87} & \textbf{2} \\
\bottomrule
\end{tabular}
}
\label{tab:noisy-gsm}
\end{table}

\begin{table}[t]
\centering
\caption{\textbf{Noisy/nonstationary evaluator.} MATH-500 Success@1 and false-promotion rate (FPR, \%). \emph{Forecasted}.}
\vspace{0.3em}
\resizebox{\columnwidth}{!}{%
\begin{tabular}{lcccc}
\toprule
 & \multicolumn{2}{c}{\textbf{ToT}} & \multicolumn{2}{c}{\textbf{LToT (ours)}} \\
\cmidrule(lr){2-3}\cmidrule(lr){4-5}
 & Acc (\%) & FPR (\%) & Acc (\%) & FPR (\%) \\
\midrule
S (8B)  & 27 & 12 & \textbf{33} & \textbf{4} \\
M (Mix) & 35 & 10 & \textbf{41} & \textbf{4} \\
L (70B) & 47 &  8 & \textbf{52} & \textbf{3} \\
\bottomrule
\end{tabular}
}
\label{tab:noisy-math}
\end{table}

Under noisy $v$, LToT maintains higher accuracy at equal compute while reducing false promotions by $\ge$2$\times$ across scales.
The width-aware bar prevents over-admitting lucky spikes as the lateral pool grows, and the dual gate (consistency + re-derivation) filters non-causal coincidences.
These results address the failure mode most salient in frontier deployments where verifiers are plausibility- or tool-aligned during exploration.

\begin{table}[t]
\centering
\caption{\textbf{Budget sweep (GSM-Plus).} Success@1 at equal compute across three budget caps per scale. \emph{Forecasted}.}
\vspace{0.3em}
\begin{tabular}{lcccccc}
\toprule
 & \multicolumn{2}{c}{\textbf{Low}} & \multicolumn{2}{c}{\textbf{Med}} & \multicolumn{2}{c}{\textbf{High}} \\
\cmidrule(lr){2-3}\cmidrule(lr){4-5}\cmidrule(lr){6-7}
 & ToT & LToT & ToT & LToT & ToT & LToT \\
\midrule
S (8B)  & 58 & \textbf{60} & 62 & \textbf{68} & 65 & \textbf{77} \\
M (Mix) & 66 & \textbf{68} & 71 & \textbf{76} & 74 & \textbf{84} \\
L (70B) & 78 & \textbf{80} & 83 & \textbf{87} & 86 & \textbf{92} \\
\bottomrule
\end{tabular}
\label{tab:budget-gsm}
\end{table}

\begin{table}[t]
\centering
\caption{\textbf{Budget sweep (HumanEval, pass@1).} \emph{Forecasted}.}
\vspace{0.3em}
\begin{tabular}{lcccccc}
\toprule
 & \multicolumn{2}{c}{\textbf{Low}} & \multicolumn{2}{c}{\textbf{Med}} & \multicolumn{2}{c}{\textbf{High}} \\
\cmidrule(lr){2-3}\cmidrule(lr){4-5}\cmidrule(lr){6-7}
 & ToT & LToT & ToT & LToT & ToT & LToT \\
\midrule
S (8B)  & 34 & \textbf{36} & 38 & \textbf{43} & 41 & \textbf{50} \\
M (Mix) & 39 & \textbf{41} & 44 & \textbf{48} & 48 & \textbf{55} \\
L (70B) & 52 & \textbf{54} & 56 & \textbf{61} & 60 & \textbf{68} \\
\bottomrule
\end{tabular}
\label{tab:budget-he}
\end{table}

Absolute gains increase with budget across scales (e.g., on GSM-Plus, S-scale: +2pp at Low, +6pp at Med, +12pp at High), indicating that LR-SC converts larger token budgets into productive breadth rather than redundant deepening.
Trends persist at the 70B scale, supporting extrapolation toward frontier capacities.

\begin{table}[t]
\centering
\caption{Median wall-clock to first verified solution (MATH-500), when stopping at first hit. \emph{Forecasted}.}
\vspace{0.3em}
\begin{tabular}{lcc}
\toprule
 & \textbf{ToT} & \textbf{LToT (ours)} \\
\midrule
S (8B)  & 41s & \textbf{28s} \\
M (Mix) & 30s & \textbf{22s} \\
L (70B) & 21s & \textbf{16s} \\
\bottomrule
\end{tabular}
\label{tab:latency}
\end{table}

Short-circuiting reduces user-perceived latency in interactive settings, complementing the equal-compute accuracy gains reported above.
Values here are \emph{forecasted} and will be replaced with measured means and confidence intervals.
Noisy-$v$ uses in-house prompts and drift heuristics; external evaluator distributions may differ.
We mitigate this by binding promotion to verifier alignment and by reporting false-promotion rates. Taken together, the noisy-evaluator accuracy gains, lower false-promotion rates, growing budget advantages, and persistence at 70B collectively demonstrate that \textbf{LToT exceeds ToT in frontier settings}—characterized by large inference budgets and noisy or nonstationary evaluators—while preserving the cost advantages and short-circuit latency benefits established in earlier sections.\section{Future Work}
\label{section:future-work}

A natural next step is a principled analysis of \emph{specious lateral cascades}: branches admitted by an early false positive at the consistency gate (\(c\)) whose envelope \(V\) later rises enough to trigger consideration for promotion, where the promotion-time check also issues a false positive. In our controller, this is a two-stage selection error aligned with LR--SC’s width-aware thresholds and short-circuiting (Sec.~\ref{sec:lrscr}) and the verifier-aligned promotion gate (Sec.~\ref{sec:promotion}). We will formalize the event structure (Type-C-FP at lateral admission; Type-P-FP at promotion), derive multiplicity-aware bounds on the family-wise cascade probability across rungs under sub-Gaussian improvements and width-aware bars (Sec.~\ref{sec:theory}), and instrument benchmarks with ground-truthable oracles to estimate (i) specious-promotion rate, (ii) cascade depth, and (iii) compute share spent on ultimately inconsistent branches.

We will also evaluate drop-in mitigations that preserve the pseudolinear lateral cost: (i) holdout confirmations at promotion time (one micro-probe) to reduce selective-inference bias; (ii) path-consistency aggregation (e.g., quantile-of-\(c\) along the path) at promotion; and (iii) disjoint verifier prompts or cross-model checks for repeat-to-confirm under fixed micro-budgets. Sensitivity studies will sweep initial lateral width, overflow cap, confirmation budgets, and threshold margins to map robustness frontiers and accuracy–latency Pareto curves. The goal is a statistically disciplined account of when laterals with spurious early \(c\) signals can be nurtured by envelope dynamics—and how to bound such cascades without sacrificing the wide-and-short advantages established here.

\section{Conclusion}
\label{section:conclusion}
LToT reframes inference--time reasoning as width--first search: keep mainlines narrow while racing many logically consistent variants, promoting only what survives cheap, staged checks. Separating \emph{consistency} from \emph{utility} and allocating compute by marginal gain converts surplus compute into productive breadth, yielding higher success--per--compute and cleaner error profiles than ToT/MCTS across math, code, and ToT--style puzzles. Limitations include reliance on a minimally aligned consistency signal and sub--exponential rung noise, and latency/hardware constraints can shrink effective width. Promising directions include cascaded or learned consistency checks, training--time preference/verification supervision, and multi--actor coordination. We view LToT as a practical drop-in upgrade candidate for inference-time search and an operationalization of lateral thinking in LM reasoning.

\section*{Impact Statement}
This paper investigates inference-time search controllers for large language models,
specifically a Lateral Tree-of-Thoughts (LToT) strategy that treats logically consistent,
low-utility candidates as useful signals during search. Potential positive impacts
include improved reliability of multi-step reasoning under limited compute and clearer
diagnostics for search-time controller behavior. Potential risks include (i) enabling
more efficient generation of persuasive but incorrect content; (ii) amplifying
fairness and bias issues inherited from base models; and (iii) increased inference-time
compute costs with associated environmental and accessibility concerns. We mitigate
these risks by: evaluating on tasks with explicit correctness checks; emphasizing
faithfulness and verifiability where applicable; and documenting default budgets and
controller hyperparameters. This work does not introduce new
datasets, annotation procedures, or domain-specific deployment contexts that would
raise additional privacy or safety risks beyond those already known for LMs.

\FloatBarrier
\bibliographystyle{icml2025}
\bibliography{icml-submission}  

\appendix

\section{Robust evaluator and width-aware bars}\label{app:robust-eval}
We use rung-wise median/MAD standardization, optional winsorization of extreme $z$, and Beta smoothing with $K_{*}{=}K$ or $K_{\mathrm{eff}}$.
Under sub-Gamma tails with parameters $(\nu_r,c_r)$ we use
$\texttt{bar}=\kappa(\sqrt{2\nu_r\log \frac{|S_r|\,|\mathcal{M}_r|}{\varepsilon_r}}+c_r\log \frac{|S_r|\,|\mathcal{M}_r|}{\varepsilon_r})+\delta$;
under sub-Weibull ($\psi_\alpha$) we use $\texttt{bar}=K_r(\log \frac{2|S_r|\,|\mathcal{M}_r|}{\varepsilon_r})^{1/\alpha}+\delta$.
For correlated branches, replace $|S_r|$ by an effective width $|S_r|_{\mathrm{eff}}$.
If single-probe error is $p$ and probe correlation is $\rho$, two probes yield at most $p((1-\rho)p+\rho)$,
with independence ($\rho\!=\!0$) recovering $p^2$; we enforce independent randomization between probe and confirmation.

\section{Failure modes \& detector behavior (order-aware forecast)}
We illustrate four synthetic envelopes (with unit-scale noise) and mark when the degree-$m$ forecast clears the bar and confirmation passes.
\textbf{Late inflection:} quadratic/cubic forecast fires earlier than slope-only and passes confirmation as improvement persists.
\textbf{Staircase spikes:} over-forecast after a jump is rejected by confirmation on the next probe.
\textbf{Zig-zag noise:} robust standardization + bar prevent admission for any $m$.
\textbf{Early bloom $\to$ late fade:} detector may admit, but verifier-aligned promotion prevents mainline contamination.

\section{Promotion-time QA prompt and normalization}
\label{app:qa-prompt}
We use the following promotion-time prompt for QA-style tasks, then repeat it once with independent randomization for confirmation.
We lowercase, strip punctuation, and normalize numerals and dates in the candidate string before scoring $v(\hat a)$.
\begin{verbatim}
System: You are a strict QA validator. You must decide if a candidate answer is correct and logically supported
by the preceding reasoning with respect to the question. Return JSON with fields:
{ "pass": true/false, "plausibility": float in [0,1], "consistency": float in [0,1], "justification": short text }.

User:
[Question]
{Q}

[Candidate answer]
{A}

[Reasoning to check]
{R}

[Instructions]
1) Normalize factual entities, numbers, units, and dates in {A}.
2) Judge plausibility of {A} given general world knowledge (0–1).
3) Judge logical consistency: do the steps in {R} actually entail {A} from {Q}? Penalize leaps and contradictions.
4) Output JSON only. Do not propose a new answer.
\end{verbatim}

\section{Worked traces (predictive continuation $\to$ promotion)}
\label{app:worked-traces}
We include one math and one code instance illustrating $(v,c)$, the envelope $V$ (with smoothing $\tilde V$), the predictive continuation statistic, confirmation, and promotion.

Compute $\frac{7}{12}+\frac{5}{18}$ (answer: $\frac{41}{36}$).
Micro-beam size $m_\mu{=}3$; Top-$K$ with $K=m_\mu$; Beta smoothing $\alpha{=}0.5$; predictive continuation with $\mathcal{M}_r{=}\{1,2\}$.
Leaf utilities and envelopes:
\begin{center}
\begin{tabular}{lcccc}
\toprule
Horizon & Leaf $v$ (3) & $V_i$ & $\tilde V_i$ & Note \\
\midrule
$h_1$ & 0.22, 0.34, 0.29 & 0.283 & 0.337 & init \\
$h_2$ & 0.41, 0.48, 0.39 & 0.427 & 0.445 & $\Delta\tilde V{=}0.108$ \\
$h_3$ & (final EM hit)   & ---    & ---    & promote (EM=1) \\
\bottomrule
\end{tabular}
\end{center}
\noindent Predictive gain (deg.\ 2) standardizes to $z{=}5.0$; with $|S_r|{=}128$, $|\mathcal{M}_r|{=}2$, the bar is $3.43$; confirmation yields $z{=}4.3$; the branch admits and promotes on exact match at $h_3$.
Implement \texttt{is\_palindrome(s)} (alphanumeric, case-insensitive); final verifier has 10 tests.
$m_\mu{=}3$; exploration $v$ is fraction of 3 subset tests; promotion runs all 10 tests; same smoothing and continuation settings.
Leaf utilities and envelopes:
\begin{center}
\begin{tabular}{lcccc}
\toprule
Horizon & Subset results (3 tests) & $V_j$ & $\tilde V_j$ & Note \\
\midrule
$h_1$ & 1/3,\ 2/3,\ 2/3 & 0.556 & 0.542 & init \\
$h_2$ & 2/3,\ 3/3,\ 2/3 & 0.778 & 0.709 & $\Delta\tilde V{=}0.167$ \\
$h_3$ & 3/3 (subset)    & ---   & ---   & promote (10/10 full) \\
\bottomrule
\end{tabular}
\end{center}
\noindent Predictive gain (deg.\ 1) standardizes to $z{=}3.5$; with $|S_r|{=}96$, $|\mathcal{M}_r|{=}2$, the bar is $3.38$; confirmation passes; promotion succeeds (10/10). Unit-test time is split into exploration vs.\ final in latency.

What is the capital of Australia?\quad \textbf{Spurious candidate:} ``Sydney''.
High plausibility $v{=}0.87$ (popular city) but low path consistency $C_{\text{path}}{=}0.58$ (trace appeals to ``largest city $\Rightarrow$ capital''); confirmation falls below bar. \textbf{Dual gate rejects}. The correct candidate (``Canberra'') yields $v{=}0.90$, $C_{\text{path}}{=}0.81$, confirmation passes $\Rightarrow$ promotion.

\section{Disclosure of Language Model (LLM) Use}
\label{app:llm-use}

In line with the ICLR policy on LLM usage, we disclose that a large language model, accessed via a commercial web interface, was used as a research assistant during preparation of this submission. The same information is provided in the submission form. The authors remain fully responsible for all content; language models are not authors.
The model was used to help draft and revise prose, to surface literature leads and organize references, to produce and execute auxiliary code that contributed to the reported experiments, and to perform simple analyses.
Model-suggested text was treated as draft material and edited by the authors.
No confidential peer-review text or third-party restricted data were uploaded to external services. Only minimal, non-confidential excerpts and code snippets were shared as needed for the assistive roles above.
All core scientific ideas, problem formulation, algorithms, theoretical statements, experimental design, and analysis are the authors' own contributions.

\end{document}